\documentclass[10pt,twocolumn,letterpaper]{article}

\usepackage{wacv}
\usepackage{booktabs}
\usepackage{times}
\usepackage{epsfig}
\usepackage{graphicx}
\usepackage{amsmath,amssymb} 
\usepackage{color}
\usepackage{amsmath}
\usepackage{amssymb}
\usepackage{subcaption}
\usepackage{enumitem}
\usepackage{xcolor}
\usepackage{setspace}
\usepackage{comment}
\usepackage{multirow}

\def\wacvPaperID{711} 

\wacvfinalcopy 

\ifwacvfinal
\def\assignedStartPage{1} 
\fi

\ifwacvfinal
\usepackage[breaklinks=true,bookmarks=false]{hyperref}
\else
\usepackage[pagebackref=true,breaklinks=true,colorlinks,bookmarks=false]{hyperref}
\fi

\ifwacvfinal
\else
\fi

\newcommand{\revv}[1]{#1}
\newcommand{\rev}[1]{#1}

\begin{document}

\title{Revisiting Edge Detection in Convolutional Neural Networks}

\author{Minh Le\\
{\tt\small minhle.r7@pm.me}
\and
Subhradeep Kayal\\
{\tt\small deep.kayal@pm.me}
}

\maketitle

\begin{abstract}
The ability to detect edges is a fundamental attribute necessary to truly capture visual concepts. In this paper, we prove that edges cannot be represented properly in the first convolutional layer of a neural network, and further show that they are poorly captured in popular neural network architectures such as VGG-16 and ResNet. 
\rev{The neural networks are found to rely on color information, which might vary in unexpected ways outside of the datasets used for their evaluation.
To improve their robustness}, we propose edge-detection units and show that they 
reduce performance loss and generate qualitatively different representations. By comparing various models, we show that the robustness of edge detection is an important factor contributing to the robustness of models against color noise.
\end{abstract}

\section{Introduction}

Among layers of a convolutional neural network, the first hidden layer has the smallest receptive field and therefore is limited to encoding the most rudimentary features.
Standard descriptions of convolutional neural networks treat it as a trivial fact that neurons in this layer \rev{can} encode edges~\cite{LeCun2015} and research effort has been spent elsewhere exploring higher-level representations, e.g.~\cite{Zeiler2014,nguyen2016}.

\begin{figure*}[!htbp]
    \newcommand{\gridsize}{0.35\textwidth}
    \centering
    \begin{subfigure}[t]{\gridsize}
        \centering
        \includegraphics[width=\textwidth]{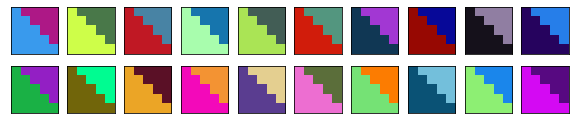}
        \caption{$45^{\circ}$ edges}
        \label{fig:45-degree-edges}
    \end{subfigure}%
    ~ 
    \begin{subfigure}[t]{\gridsize}
        \centering
        \includegraphics[width=\textwidth]{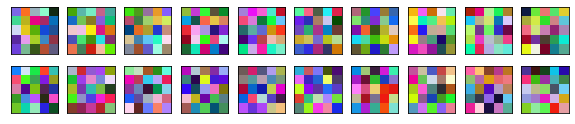}
        \caption{Uniform noise}
        \label{fig:noise}
    \end{subfigure}
    \caption{Examples of edges and uniform noise patterns. In our experiments, a very simple neural network needs to classify patches like these into the right category. Best view in color.}
    \label{fig:patches}
\end{figure*}

However, an edge is a non-trivial concept. Consider as a case study the concept \textit{45$^\circ$ edges from the top-left corner to the bottom-right corner of a 5 $\times$ 5 patch} (Figure~\ref{fig:45-degree-edges}). With a reasonable bit depth, millions examples of this concept can be generated, each with a distinct appearance.
This complexity is in stark contrast with the simplicity of an artificial neuron. Given that the net input of a unit in the first layer is a linear combination of RGB intensities, it is likely that it will learn a combination of an edge and certain colors instead of a generalized notion of an edge. 

While colors are undoubtedly an important type of information that can help models perform well on independent, identical distributed (i.i.d.) datasets, relying on colors for object recognition can also limit the ability of models to capture the essence of concepts. 
\rev{The color of an object varies in response to lighting condition and the distribution of colors varies with seasons, geography, or societal context. A robust computer vision system should perform equally well across such conditions~\cite{Marcus2020}.
With the increasingly common presence of AI in everyday life, it is more and more important that robustness is guaranteed. The practical consequence of failure might be, for example, accidents caused by autopilot systems that recognize pedestrians at noon but fail to do so in twilight or recognize yellow trucks but not white ones.}

In this paper, we hypothesize that:
\begin{enumerate}[itemsep=-0.5ex]
\item Standard artificial neurons cannot learn the concept of an edge independent of colors, and
\item The conflation of shapes and colors is one of the factors underlying the inability to handle negative images~\cite{Hosseini2017} and color-shifting~\cite{Hosseini.poovendran18}.
\end{enumerate}{}

We study edge detection in neural networks using two methods.
First, we use synthetic stimuli, inspired by research in primate visual systems \cite{Hubel.wiesel1968}, to test whether individual neurons of pre-trained state-of-the-art neural networks have learned to discriminate edges.
Second, we design neurons that enforce theoretical requirements of edge detection and contrast their behaviors with that of standard neural networks.
Experiments are conducted on \mbox{CIFAR-10~\cite{krizhevsky09}} and a scaled-down version of ImageNet~\cite{deng09.imagenet} using two popular architectures in computer vision: VGG~\cite{Simonyan.zisserman2015} and ResNet~\cite{He16resnet}.
Our contributions are threefold:
\begin{enumerate}[itemsep=-0.5ex]
    \item We show \rev{evidence supporting the hypothesis that standard convolutional neural networks do not reliably capture edges independent of color.}
    \item We show that poor performance on negative and color-shifted images can be partly explained by a failure to learn color-independent edges.
    \item We propose edge detection neurons and show that neural networks equipped with them perform more robustly against color noise. \rev{We also show that most of the performance of CNNs can be achieved without access to absolute color information.}
\end{enumerate}

{}

Edge detection is a task and a method of feature extraction. In this paper, instead of producing edge-maps such as in \cite{Akbarinia.parraga2018,Soria2019}, we will focus on how to extract robust features for image recognition.
To the best of our knowledge, we are the first to design artificial neurons specifically to capture edges.
In image recognition, alternative convolutional layers have been proposed such as local binary convolution~\cite{Juefei-Xu2017}
and quadratic convolution~\cite{Zoumpourlis17}.
Gabor filters were used in~\cite{Chen2017} to replace the first convolutional layer to reduce sample complexity.
In the broader literature on neural networks, \cite{Madsen.johansen2020}~proposes specialized units to perform arithmetic operations.

The paper is structured as follows: Section~\ref{sec:synthetic} presents theoretical and empirical results on the ability of standard neurons to represent edges. Section~\ref{sec:edge-detector} describes our proposal of edge-detecting neurons. Section~\ref{sec:exp-settings} and Section~\ref{sec:results} detail the settings and results of our experiments and, finally, Section~\ref{sec:conclusions} offers some concluding remarks.

\section{Edge Detection on Synthetic Data}
\label{sec:synthetic}

To test the ability of models on learning color-independent edges, we formulate a binary classification task. Inputs are trichromatic images of the size $5 \times 5$ with intensities in the range of [0,~1].
The positive class contains 45$^\circ$ edges in which the left half and the right half are different in at least two channels by at least~$\epsilon$. We find that $\epsilon=0.4$ makes the edges clearly visible (Figure~\ref{fig:45-degree-edges}).
The negative class contains uniform noise as depicted in Figure~\ref{fig:noise}.
Although the task is simple for humans, we will show that it is not linearly separable and pose significant difficulty to standard artificial neurons.

\subsection{Linear Inseparability}

\newcommand{\rff}{\mathbb{R}^{5 \times 5}}
\newcommand{\rgb}{{\{r,g,b\}}}

The problem can be defined as classifying a tuple of matrices $(r,g,b)$, where $r,g,b \in \rff$, into one of the two classes defined above. 
Let $L$ be the indices of all pixels in the left part of the edge and $R$ those in the right, the differences between the average of intensities between the two halves are:
$\Delta_r = \overline{r_\mathrm{R}}- \overline{r_\mathrm{L}}$,
$\Delta_g = \overline{g_\mathrm{R}}- \overline{g_\mathrm{L}}$, and
$\Delta_b = \overline{b_\mathrm{R}}- \overline{b_\mathrm{L}}$.
\revv{If we reparametrize in terms of $\Delta_{\{r,g,b\}}$,} the problem is reduced to a binary classification of 3-dimensional vectors. We will show that this problem is not linearly separable.

\revv{The positive class is limited to a region of the space that has $\left(\left|\Delta_r\right| \ge \epsilon\right) \wedge \left(\left|\Delta_g\right| \ge \epsilon\right) \wedge \left(\left|\Delta_b\right| \ge \epsilon\right)$ (but do not fill this region).}
Because the negative class consists of uniform noise, using the properties of expectation and standard deviation, we can establish that $\Delta_{\{r,g,b\}}$ are distributed according to $\mathcal{N}(0, \sigma)$ with $\sigma \approx 0.12$.
Therefore, for $\epsilon=0.4$, \revv{most of} the negative class \revv{lies within the range $[-\epsilon; \epsilon]$ along each dimension:}
$\mathbb{P}(\left|\Delta_r\right| < \epsilon) = \mathbb{P}(\left|\Delta_g\right| < \epsilon) = \mathbb{P}(\left|\Delta_b\right| < \epsilon) \approx 0.999$.
In geometric terms, the negative class \revv{has more} than 99.7\% of its mass inside \revv{a box while a small portion of the negative class together with the positive class are scattered outside of the box in every direction}. It is easy to see that this is a linearly inseparable problem and therefore cannot be solved by a standard neuron.
Similar results hold for smaller values of~$\epsilon$.

\revv{If we do not use the above parametrization, the problem is not linearly separable either.
Each point in the $\Delta_\rgb$ space corresponds to a subspace of the input space. Consider a linear classifier with weights that cannot be re-expressed by $\Delta_\rgb$ and a point $P$ of class $C$ which specifies the subspace $\mathcal{S}$ in the input space. Because the weights cannot be rewritten as $\Delta_\rgb$, the classifier hyperplane intersects with the subspace $\mathcal{S}$ and cuts it in halves, both belong to the class $C$. In other words, any linear decision boundary is guaranteed to have points of the same class on both of its sides.}

\subsection{The (In)ability of Standard Neurons to Represent Edges}
\label{ssec:patch-experiments}

\begin{table*}[t]
    \centering \small
\begin{tabular}{r@{\hskip 1em}l@{\hskip 1em}c@{\hskip 1em}c@{\hskip 1em}c}
\hline
& Model  &    $n=100$ &    $n=500$ &    $n=1000$ \\
\hline
1. & Standard unit & 0.509 (0.090) &  0.476 (0.092) &  0.502 (0.073) \\
2. & Layered, $h=1$ &  0.829 (0.144) &  0.920 (0.099) &  0.920 (0.113) \\
3. & Layered, $h=2$ &  0.884 (0.113) &  0.980 (0.040) &  0.985 (0.038) \\
4. & Layered, $h=3$ & 0.986 (0.028) &  0.994 (0.015) &  0.999 (0.006) \\
5. & Edge detector unit & 0.991 (0.017) &  0.999 (0.006) &  0.999 (0.006) \\
\hline
\end{tabular}
    \caption{Performance of different neural architectures on classifying edges versus random noise. $h$ is the number of nodes in the hidden layer. Reported figures are accuracy after $n$ parameter updates.}
    \label{tab:patch-experiment-results}
\end{table*}{}

To confirm the theoretical insight, we perform an experiment where a single neuron is trained to classify images generated according to the specifications in the previous section.
1000 iterations were carried out; in each of them, 16 examples of edges and 16 noise patterns are generated and used to perform one update of the neuron's parameters. The accuracy of the neuron at the $n$\textsuperscript{th} iteration is reported for $n \in \{ 100, 500, 1000\}$.
The experiment is repeated 25 times to get an estimate of variation.
The first row in Table~\ref{tab:patch-experiment-results} demonstrates that the standard neuron cannot capture edges correctly. 

Existing literature suggests that adding a hidden layer can overcome linear inseparability.
Therefore, we experimented with a minimized version of VGG~\cite{Simonyan.zisserman2015} which contains a convolutional layer with $h$ filters of size $3 \times 3$ followed by batch normalization, ReLU activation, and a dense output layer. 
The results in rows (2-4) indicate that $h=3$ is enough to robustly capture edges. In other words, a neuron in the second layer of a VGG network can perform edge detection by combining the output of at least 3 neurons in the first layer. 

\subsection{Edge Detection in Neural Networks Trained on Natural Images}
\label{sec:edge-test}

\begin{figure*}[bhtp]
    \centering
    \begin{subfigure}[t]{0.3\textwidth}
        \centering
        \includegraphics[width=0.93\textwidth]{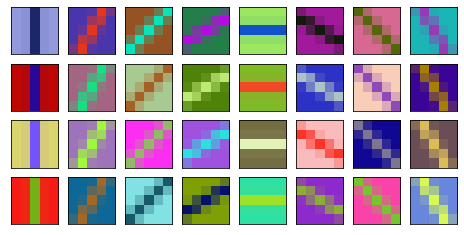} 
        \caption{Some edge stimuli generated with random colors}
        \label{fig:stimuli}
    \end{subfigure}%
    ~ 
    \begin{subfigure}[t]{0.23\textwidth}
        \centering
        \includegraphics[width=\textwidth]{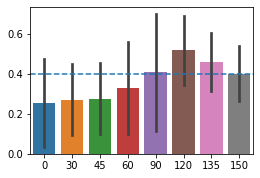}
        \caption{The activation of an orientation-selective neuron}
        \label{fig:orientation-selective}
    \end{subfigure}%
    ~ 
    \begin{subfigure}[t]{0.23\textwidth}
        \centering
        \includegraphics[width=\textwidth]{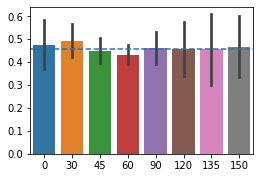}
        \caption{The activation of a neuron without clear preference for an edge}
        \label{fig:non-selective}
    \end{subfigure}%
    \caption{Some examples of the stimuli used in our experiment (a) and the activation of randomly selected neurons (b,c) on edges oriented in different angles. Horizontal dashed lines show activation on uniform noise and error bars represents standard deviation.}
    \label{fig:my_label}
\end{figure*}

Having enough capacity is a necessary but not sufficient condition for learning a concept. It is doubtful that neurons would spontaneously learn edges when trained on natural images because they would observe much less variation and the supervisory signal would be less explicit.
We therefore evaluate neurons of a pretrained VGG-16 network (accuracy on ImageNet: 73.4\% top-1, 91.5\% \mbox{top-5}) on edge-versus-noise classification. We extend the previous experiment to include edges of round degrees between $0^\circ$ and $180^\circ$ (Figure~\ref{fig:stimuli}), similar to the paradigm used to study edge detection in neuroscience \cite{Hubel.wiesel1968}. 
For each angle, 10,000 images are randomly generated, coupled with the same number of random noise samples. A neuron is said to predict an edge if its activation is higher than a certain threshold. The optimal threshold is determined for each neuron based on test data so that we could calculate the upper-bound of accuracy.
We report the maximum accuracy among neurons of each of the first five convolutional layers in the studied neural network. 
The higher a layer is, the less meaningful the test becomes because the receptive field gets much bigger than the stimuli.

From Table~\ref{tab:edge-detection-real-neurons}, it is immediately clear that neurons in the second convolutional layer do not spontaneously learn to represent edges. From layer 3 onward, \rev{we can often find at least one neuron for each angle that distinguish between edges and noise with an accuracy of at least 90\% (with an exception for 45$^\circ$ edges for which the accuracy maxes out at 87\%).} 
Interestingly, we found in layer 3 neurons that activate for edges of a certain orientation (Figure~\ref{fig:orientation-selective}) similar to what has been observed in cats and monkeys \cite{hubel1959receptive,Hubel.wiesel1968}. However, we also find many neurons that do not show a preference for any particular orientation (Figure~\ref{fig:non-selective}). 
Overall, 60.9\% of neurons can predict edges of a certain orientation better than chance (50\%) and only 18.0\% can perform at 75\% accuracy or higher.
Moreover, among the neurons that perform better than chance, we find that the coefficient of variation is greater than 1 in 14.9\% of cases and greater than 1.5 in 5.8\%. 

The observations in this section suggest that the representations of edges in VGG-16 trained on ImageNet are dependent on colors.
Therefore, to explore the possibility of robust edge detection in neural networks, we propose a theoretically grounded design of edge detection neurons.

\newcommand{\e}[1]{\raisebox{-1.1ex}{\includegraphics[width=1.5em]{images/edge#1.png}}$#1^\circ$}

\begin{table*}[t]
\small
\centering

\begin{tabular}{c@{\hskip 1em}c@{\hskip 1em}c@{\hskip 0.8em}c@{\hskip 0.8em}c@{\hskip 0.8em}c@{\hskip 0.8em}c@{\hskip 0.5em}c@{\hskip 0.45em}c}
\hline 
Conv.\ layer & \e{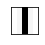} & \e{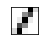} & \e{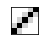} & \e{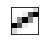} & \e{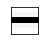} & \e{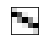} & \e{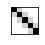} & \e{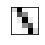} \\
\hline
     1 ($3 \times 3$) & 0.563 & 0.559 & 0.559 & 0.557 & 0.559 & 0.558 & 0.558 & 0.560 \\
     2 ($5 \times 5$) & 0.626 & 0.607 & 0.607 & 0.610 & 0.605 & 0.602 & 0.601 & 0.600 \\
     3 ($9 \times 9$) & 0.977 & 0.924 & 0.871 & 0.963 & 0.981 & 0.938 & 0.928 & 0.928 \\
     4 ($11 \times 11$) & 0.968 & 0.931 & 0.820 & 0.884 & 0.875 & 0.941 & 0.960 & 0.963 \\
     5 ($15 \times 15$) & 0.887 & 0.877 & 0.822 & 0.893 & 0.925 & 0.808 & 0.809 & 0.825 \\
\hline
\end{tabular}

    \caption{The upper bound of the accuracy in edge-vs-noise classification for the first convolutional layers of VGG-16 trained on CIFAR-10. ``Conv.\ layer'' column specifies the number of the convolutional layer and the size of its receptive field.}
    \label{tab:edge-detection-real-neurons}
\end{table*}

\section{Edge Detection Neurons}
\label{sec:edge-detector}

\rev{An edge is defined as} ``a location of rapid intensity variation'' \cite[p.\ 211]{Szeliski2011}. What are missing from this definition, and hence edge detectors should be invariant to, are the absolute value of intensity, what color channel is considered, and the direction of change.
We therefore propose edge-detecting neurons with \rev{three characteristics. First, patches are mean-centered such that only relative differences matter. Second, weights for different input channels differ by a scalar multiplier only so that the same form is captured. Finally, the absolute of channel-specific outputs are summed so that the direction does not matter and, importantly, they do not cancel out each other.}
\rev{The activation of an edge-detecting neuron at location $(x,y)$ is therefore given by:
\begin{equation}
    o_{x,y} = \sum_{c \in \mathcal{C}} \alpha_{c} | w \cdot (p_{x,y,c} - \overline{p_{x,y,c}})| + b,
\label{eq:edge-detection}
\end{equation}
where $o_{x,y} \in \mathbb{R}$ is the activation at the specified location, $c$ is an input channel, $w \in \mathbb{R}^{n \times n}$ is a weight matrix shared across input channels, \rev{a dot ($\cdot$) denotes the dot product of flattened matrices}, $p_{x,y,c} \in \mathbb{R}^{n \times n}$ is a single-channel patch around  the target pixel, an overline ($\overline{x}$) denotes the mean, $\alpha_{c} \in \mathbb{R}^+$ is a channel-\rev{specific} weight, and $b \in \mathbb{R}$ is the bias term.}

\rev{To gain more insight, let us define channel-specific term $o_{x,y,c}$ such that $o_{x,y} = \sum_c \alpha_c o_{x,y,c} + b$. The term can be rewritten (via expanding and regrouping) as:
\begin{equation}
    o_{x,y,c} = |w \cdot (p_{x,y,c} - \overline{p_{x,y,c}})| = |(w-\overline{w}) \cdot p_{x,y,c}|
\end{equation}

In words, the proposed model enforces a constraint of zero mean on kernels. This is compatible with popular edge detecting operators such as Roberts cross, Prewitt \cite{prewitt1970}, and Sobel. 
The kernel of our neuron is learned and, therefore, dependent on the task at hand, possibly taking novel shapes.}

Before testing the design on natural images, we evaluate it on the task of learning $45^\circ$ edges as described in the previous section. Row~5 in Table~\ref{tab:patch-experiment-results} shows that edge detection neurons converge faster and achieve higher levels of accuracy than conventional neurons.

To incorporate edge detection neurons into a neural network, one can simply replace the first linear transformation with the transformation specified in Equation~\ref{eq:edge-detection}. Because this will result in a loss of absolute color information, 
we can expect model performance to degrade. However, in the current paper, we focus on studying the properties of neural networks as opposed to optimizing for performance. It is also of scientific interest to know \textit{how much} the performance degrades because it shows the relative usefulness of edge and color information.

\section{Experimental Settings}
\label{sec:exp-settings}

To study edge detection in neural networks, we will compare models with different performance profiles (i.e. performance on natural, negative, and color-shifted images) and contrast models with edge-detecting neurons and regular models.
Three components are needed for the experiments: datasets, transformations, and models. We will describe each of them in the remainder of this section.

\newcommand{\inlinesec}[1]{\vspace{0.8em}\textbf{#1}}

\subsection{Datasets}

\vspace{-0.8em}
\inlinesec{CIFAR-10} \cite{krizhevsky09} contains $32 \times 32$ color images of 10 types of objects in natural settings. There are 50,000 training and 10,000 testing examples.

\inlinesec{Tiny ImageNet}\footnotemark{} is a scaled-down version the popular ImageNet dataset \cite{deng09.imagenet}. It contains $64 \times 64$ images in 200 classes. The training set contains 100,000 training examples equally divided into classes. The validation and test set contain 10,000 examples each. Because labels are not provided for test examples, we split the validation set into halves and use one for development, the other for testing.
\footnotetext{\url{https://tiny-imagenet.herokuapp.com/}}
 
\subsection{Transformations}
\label{sec:transformations}

\begin{figure*}[thbp]
    \newcommand{\gridsize}{0.28\textwidth}
    \begin{subfigure}[t]{\gridsize}
        \centering
        \includegraphics[width=\textwidth]{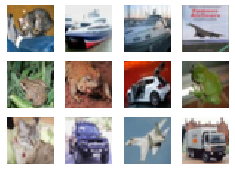}
        \caption{Original images}
        \label{fig:cifar-original}
    \end{subfigure}%
    ~ 
    \begin{subfigure}[t]{\gridsize}
        \centering
        \includegraphics[width=\textwidth]{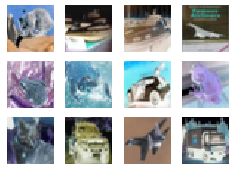}
        \caption{Negative images}
        \label{fig:cifar-negative}
    \end{subfigure}%
    ~ 
    \centering
    \begin{subfigure}[t]{\gridsize}
        \centering
        \includegraphics[width=\textwidth]{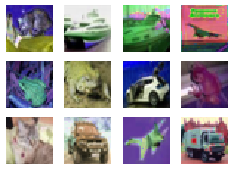}
        \caption{Color-shifted images}
        \label{fig:cifar-recolored}
    \end{subfigure}%
    \caption{Examples of original CIFAR-10 images and corresponding transformed images.}
    \label{fig:cifar10}
\end{figure*}{}

\rev{Following the literature on CNN robustness (e.g.\ \cite{Qian2019}, \cite{Ghosh2019}), we evaluate models on image classification.}
It has been demonstrated that the performance of neural networks drops sharply when images are recolored while humans see virtually no performance loss \cite{Hosseini2017}. We will study two kinds of \rev{color-related} transformations reported in the literature (Figure~\ref{fig:cifar10}).

\inlinesec{Negative images} \cite{Hosseini2017} cause substantial performance loss.
The effect is consistent across model architectures and datasets. The authors showed that it is possible to increase the test score on negative images by sacrificing the performance on regular images. 

\inlinesec{Color-shifting} \cite{Hosseini.poovendran18}
is proposed as a method of adversarial perturbation. 
Similar to the case of negative images, the authors observe that it is possible to increase the robustness of models on color-shifted images by sacrificing some performance on natural images. 
We adopt the transformations defined by \cite{Hosseini.poovendran18} to test the robustness of models against color noise but do not address the adversarial setting. Another difference is that we keep the noise added to the saturation channel small enough to avoid distorting the structure of images.


\subsection{Models}

\vspace{-0.8em}
\inlinesec{VGG}~\cite{Simonyan.zisserman2015} is a model that uses small kernels stacked in many layers to achieve high performance. We adopt VGG with a bigger kernel in the first layer to facilitate visualization and lower the number of layers to speed up training.
The modified architecture achieves around 83\% accuracy on \mbox{CIFAR-10}.
All models are trained for 75 epochs and the epoch that achieves the best performance on the validation set is kept.

\inlinesec{ResNet}~\cite{He16resnet} introduces residual connections to speed up convergence. The model and its variations consistently achieve high performance on ImageNet. We adopt ResNet-18 for our experiments on Tiny ImageNet, with some modifications to accommodate smaller images. 
All models are trained for 20 epochs, keeping the epoch with the best performance on the validation set.

\inlinesec{}
Models are trained 10 times with different random seeds to estimate the distribution of performance. Where applicable, color-shifting augmentation is applied randomly to training images 
using the transformation described in Section~\ref{sec:transformations}. Other augmentation methods such as cropping, rotating, and flipping are the same across all models.

\section{Results and Discussions}
\label{sec:results}

In this section we will discuss the results of our experiments and the insights drawn from them.

\rev{\subsection{Robustness}

We measure robustness by the relative difference between the performance on transformed images and natural ones. This is denoted by $\Delta$ in Table~\ref{tab:test-results}. All figures are negative because models suffers from performance loss.}

Adding edge-detection neurons makes a big difference in the robustness of models, coupled with the expected sacrifice in performance on natural images. \rev{For example, performance loss is reduced from 25.8\% (Row~1) to 5.6\% (Row~2) in the case of CIFAR-10 images under color shift}. The difference is more prominent in the case of negative images, where the performance losses are reduced from 60.5\% (Row~1) to only 1.6\% (Row~2) for CIFAR-10 images, and from 77.9\% (Row~5) to 3.7\% (Row~6) for Tiny ImageNet. Negative images represent a drastic change to the color cue and therefore it is understandable that our models, which use edge features, fare much better than a conventional model, which might learn to rely to large extent on color.

\rev{Data augmentation is more efficient than edge detection neurons in increasing robustness against color-shifting (the same transformation used to generate augmented examples) and less effective for negative images. Combining data augmentation and inductive bias, however, produces the best result. This might reflect the fact that the proposed edge detection units only modify the first layer while data augmentation affects the whole network.}

\subsection{Explaining the Performance on Negative and Color-shifted Images}

\rev{In addition to reporting summary performance, we are interested in locating the source of performance and identifying attributes of neurons that contribute to the overall robustness of the model.
To this end, we measure} the accuracy on edge-vs-noise classification and the coefficient of variation on randomly generated edges (Figure~\ref{fig:stimuli}). 
Measurements are done on the edge-detection layer for \rev{applicable} models and the second convolutional layer for \rev{the rest}.\footnotemark{}
The results can be found in Table~\ref{tab:test-results}. 
\footnotetext{This layer has a receptive field of the size $9 \times 9$ for both VGG and ResNet models. We do not test higher layers because their receptive field exceeds the size of the test patterns too much.}

\begin{table*}[t]
\newcommand{\s}{\hskip 0.5em}
\newcommand{\cols}{\hskip 0.5em}
\newcommand{\mystretch}{\setstretch{0.8}}
\newcommand{\ms}[2]{\parbox[t]{4em}{\mystretch \centering #1\\{\footnotesize \color{gray} (#2)}}}
\newcommand{\msp}[4]{\parbox[t]{6em}{\mystretch \centering #1/#3\%\\{\footnotesize \color{gray} (#2/#4\%)}}}
    \small \centering
    \begin{tabular}{lll@{\hskip 0.3em}c@{\s}c@{\s}c@{\cols}c@{\hskip 1em}c}
    \toprule
    & Dataset & Model & Edge acc. & Edge var. & Regular & Negative/$\Delta$  & Color/$\Delta$  \\
    \midrule
1. & CIFAR-10 & VGG       &  \ms{0.848}{0.024} & \ms{2.640}{0.324} & \ms{0.828}{0.003} & \msp{0.327}{0.017}{-60.5}{2.1} &  \msp{0.614}{0.015}{-25.8}{1.7} \\
2. & CIFAR-10 & VGG+edge & \ms{0.869}{0.022} & \ms{1.382}{0.127} & \ms{0.775}{0.013} & \msp{0.763}{0.012}{-1.6}{0.3} &  \msp{0.732}{0.011}{-5.6}{0.5} \\
3. & CIFAR-10 & VGG+aug   &  \ms{0.940}{0.009} & \ms{2.539}{0.134} & \ms{0.823}{0.005} & \msp{0.513}{0.014}{-37.7}{1.6} &  \msp{0.805}{0.005}{-2.3}{0.3} \\
4. & CIFAR-10 & VGG+aug+edge &  \ms{0.868}{0.019} & \ms{1.455}{0.087} & \ms{0.811}{0.004} & \msp{0.811}{0.004}{-0.1}{0.1} &  \msp{0.804}{0.003}{-0.9}{0.2} \\
    \midrule
5. & Tiny ImageNet & ResNet & \ms{0.869}{0.019} & \ms{1.466}{0.047} & \ms{0.524}{0.008} & \msp{0.116}{0.007}{-77.9}{1.3} &   \msp{0.310}{0.013}{-41.0}{2.3} \\
6. & Tiny ImageNet & ResNet+edge  & \ms{0.813}{0.013} & \ms{1.236}{0.099} & \ms{0.441}{0.007} & \msp{0.425}{0.007}{-3.7}{0.9} &   \msp{0.336}{0.010}{-23.9}{1.8} \\
7. & Tiny ImageNet & ResNet+aug & \ms{0.881}{0.011} & \ms{1.412}{0.150} & \ms{0.486}{0.005} & \msp{0.197}{0.007}{-59.6}{1.2} &  \msp{0.467}{0.006}{-3.9}{0.5} \\
8. & Tiny ImageNet & ResNet+aug+edge & \ms{0.842}{0.022} & \ms{1.127}{0.052} & \ms{0.423}{0.009} & \msp{0.416}{0.009}{-1.8}{0.9} &   \msp{0.412}{0.009}{-2.6}{0.2} \\
    \bottomrule
    \end{tabular}
    \caption{Performance of models in different settings. Measurements: ``Edge acc.''=accuracy on edge-vs-noise classification, ``Edge var.''=coefficient of variation for edges of random colors, 
    ``Regular''=original test set, ``Negative''=the test set turned into negative images, ``Color''=color-shifted test set, ``$\Delta$''=relative performance change between transformed and original datasets. Models: ``+edge'' means the first convolutional layer is replaced with a layer of edge-detecting neurons, ``+aug'' means models are trained with color-shift augmentation. Figures are in the form of mean (std.) as calculated on 10 trials.}
    \label{tab:test-results}
\end{table*}

\rev{Our edge detection layers do not always recognize more edges than the second layer of regular networks (``Edge acc.'' is lower or shows no statistically significant diffence.
However, the activation of the edge detection layers is consistently more stable, reflected in lower ``Edge var.''}

\rev{Both ``Edge acc.'' and ``Edge var.'' show mild correlation with $\Delta_\mathrm{Color}$. The magnitude of Pearson's $\rho$ is around 0.57 for CIFAR-10 and 0.20 for Tiny ImageNet. When compared to $\Delta_\mathrm{Negative}$, ``Edge acc.'' sends mixed signals while ``Edge var.'' shows highly negative correlation (-0.96 for CIFAR-10 and -0.99 for Tiny ImageNet), meaning smaller variation is a good predictor of robustness against negative images.}
This result shows that increasing the robustness of edge detection is a viable way to improve the overall robustness of models.

Why is the performance drop of ``+edge'' networks much lower on negative images compared to that of color-shifted images while the former are apparently changed in a more radical way? We hypothesize that because edge detection units use \rev{unsigned difference}, it is particularly robust against color inversion. To validate this hypothesis, we calculate the normalized change in the activation edge detection units as follows:

\renewcommand{\neg}{\mathrm{negative}}
\newcommand{\reg}{\mathrm{regular}}
\begin{equation}
    \Delta_{\neg} = \frac{E\left( \left| a_\neg - a_\reg \right|_1 \right)}{E\left( \left| a_\reg \right|_1 \right)},
\end{equation}
where $a_\reg, a_\neg \in \mathbb{R}^m$ are the activation of $m$ edge detection neurons on a regular or negative image, and $E(x)$ is the mean of $x$.

For CIFAR-10, $\Delta_{\mathrm{negative}} \approx 10^{-7}$, proving that edge detection units block the perturbation from reaching higher layers. For Tiny ImageNet, $\Delta_{\mathrm{negative}} = 0.299$ (std=0.007) without augmentation and $\Delta_{\mathrm{negative}} = 0.253$ (std=0.008) with augmentation, which are substantially lower than the variation on color-shifted images. 
They are also much lower than the corresponding figures measured on the second layer of regular networks ($\Delta_{\mathrm{negative}} = 0.717$, std=0.072 with normal training and $\Delta_{\mathrm{negative}} = 0.613$, std=0.056 with color-shift augmentation).

\begin{figure*}[!htbp]
    \newcommand{\gridsize}{0.35\textwidth}
    \begin{subfigure}[t]{\gridsize}
        \centering
        \includegraphics[width=0.95\textwidth]{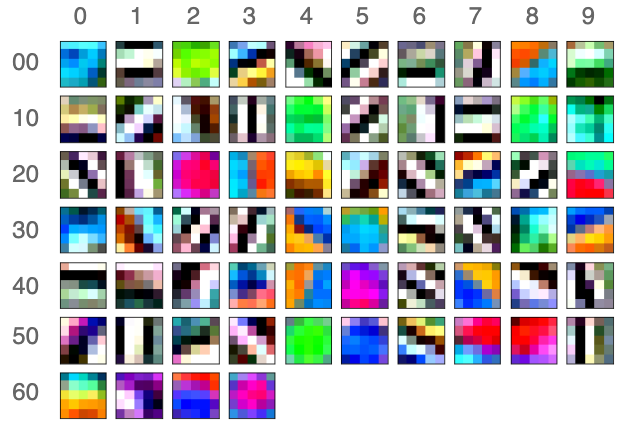}
        \caption{Regular neurons}
        \label{fig:weight-regular}
    \end{subfigure}%
    ~ 
    \centering
    \begin{subfigure}[t]{\gridsize}
        \centering
        \includegraphics[width=0.95\textwidth]{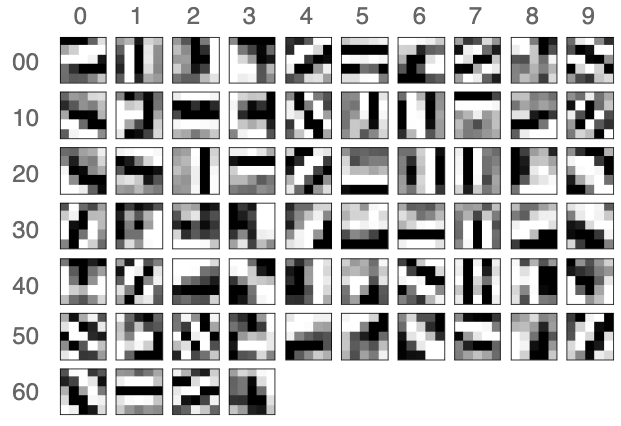}
        \caption{Edge detectors}
        \label{fig:weight-edge}
    \end{subfigure}%
    \caption{Weights of units taken from the first layer of a regular neural network and one using edge detectors.}
    \label{fig:weight-viz}
\end{figure*}{}

Finally, it has been reported in \cite{Hosseini.poovendran18} that data augmentation improves the performance on color-shifted images but how this operation changes the latent representations of a neural network is not clear. The results of our experiments indicate that the augmentation \rev{facilitates the formation of edge detectors}, reflected in higher ``Edge acc.'', while having little effect on the robustness of their detection, reflected in small differences in ``Edge var.'' compared to the standard deviation.
Comparing edge-detection-enabled models with and without augmentation reveals that a big part of what augmentation does is to train higher layers to cope with variation passed through the first layers: while ``Edge acc.'' and ``Edge var.'' change little, the performance on color-shifted images improves significantly.
This result suggests that while data augmentation often improves performance, it might not improve the ability to capture abstract concepts. Instead, deep neural networks use their vast capacity to figure out other non-intuitive ways to solve the dataset.

\subsection{Weight Visualization}

Previous sections have studied the effect of edge detection quantitatively. In this section, we will zoom in to the proposed edge detection neurons to see how they qualitatively change the internal representations.
Figure~\ref{fig:weight-viz} plots the weights of all neurons in the first layer of a regular network and one with edge detectors. Both models are trained on CIFAR-10.
For the edge detectors, because the normalized channel weights have an average of 0.333 (std=0.034), i.e.\ the 3 channels contribute approximately equally, we plot them in gray scale for better viewing.

\rev{In both models, we can find neurons that resemble edge, ridge, corner, or blob detectors. However, there are also big differences.}
A prominent difference is that Figure~\ref{fig:weight-regular} features various patterns of a circle surrounded by a darker color (neurons 2, 14, 18, 22, 35, 45, 54, and 63) whereas Figure~\ref{fig:weight-edge} has only two patterns that fit this description to some extent (neurons 29 and 51). The regular neurons count among themselves a unit for $45^\circ$-edge with the top half being deep orange and the other light blue (neuron 8), one for the same edge but with light orange-deep blue transition (neuron 44). They also contain neurons for $135^\circ$ edges (34 and 47) with opposite transitions. In contrast, Figure~\ref{fig:weight-edge} contains only one copy for each type of shapes.

In place of colors, the capacity of edge detectors is reserved for different shapes. Detected patterns include some of the edges discussed in the previous sections. We can also identify single edges of different orientations (neurons 33, 35, 39, 42, 54), a thin line (neurons 0, 18, 20, 29, 34), multiple lines (neurons 4, 5, 7, 9, 10, 50, 52, etc.), a line on a circle (1, 61), and corners (neurons 2, 17, 63).

\begin{figure*}[thbp]
    \newcommand{\gridsize}{0.24\textwidth}
    \begin{subfigure}[t]{\gridsize}
        \centering
        \includegraphics[width=\textwidth]{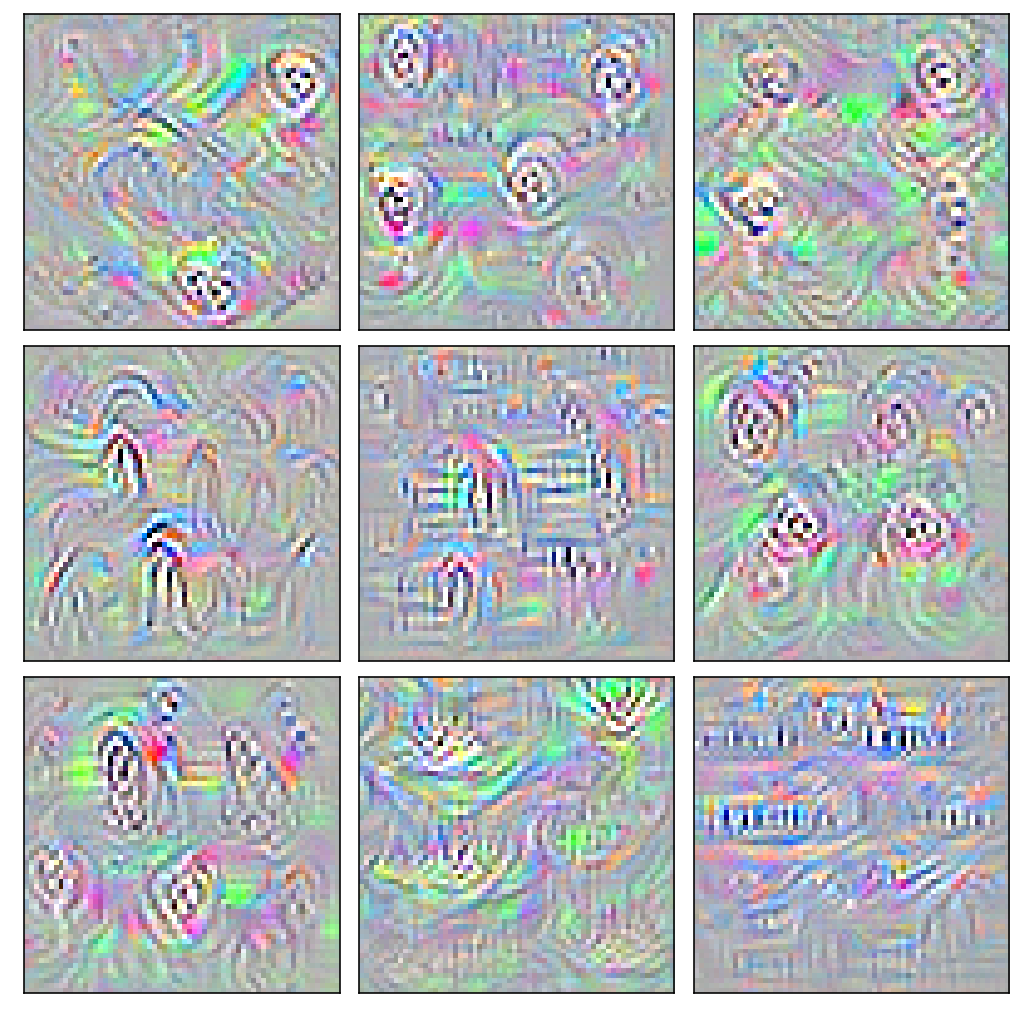}
        \caption{VGG, CIFAR-10}
        \label{fig:max-activation-regular}
    \end{subfigure}%
    ~ 
    \centering
   \begin{subfigure}[t]{\gridsize}
        \centering
        \includegraphics[width=\textwidth]{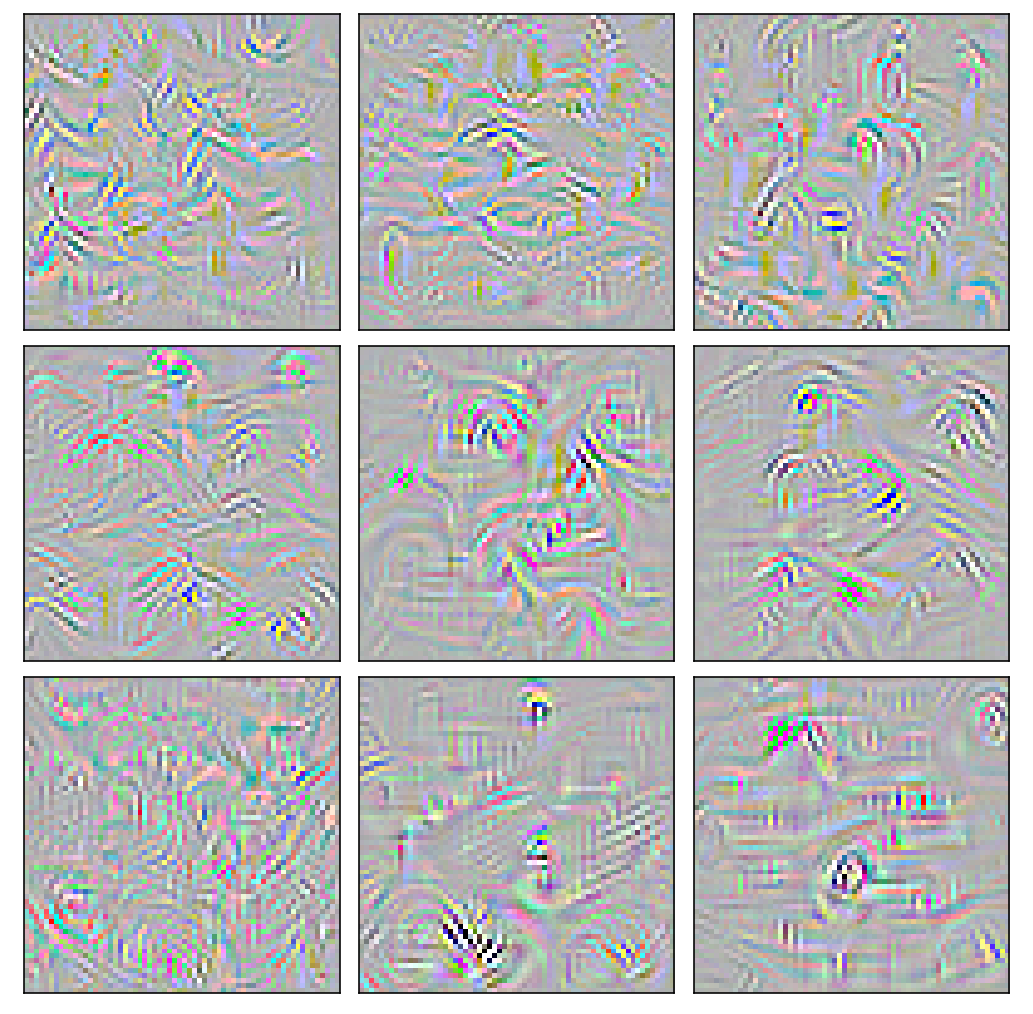}
        \caption{VGG+edge, CIFAR-10}
        \label{fig:max-activation-edge}
    \end{subfigure}
    ~
    \begin{subfigure}[t]{\gridsize}
        \centering
        \includegraphics[width=\textwidth]{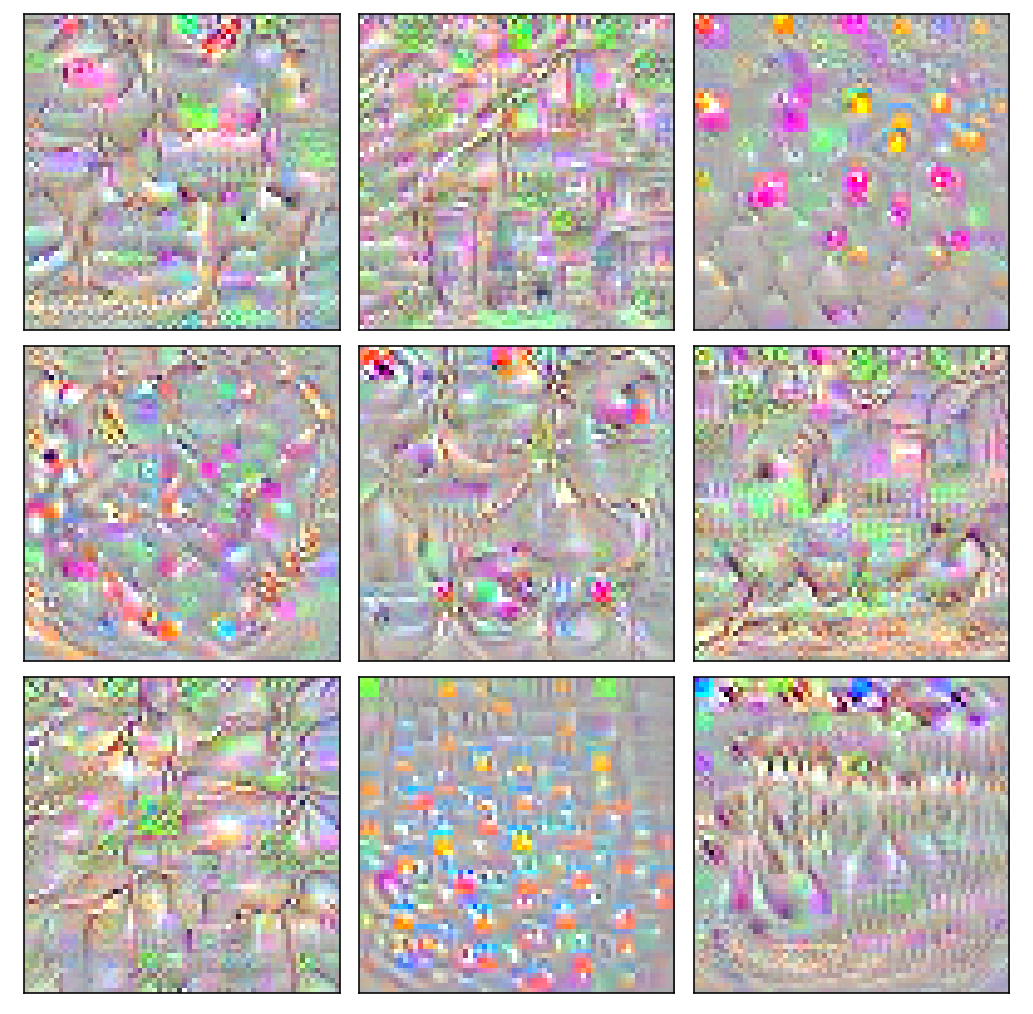}
        \caption{ResNet, Tiny ImageNet}
        \label{fig:max-activation-regular-imagenet}
    \end{subfigure}%
    ~ 
    \centering
    \begin{subfigure}[t]{\gridsize}
        \centering
        \includegraphics[width=\textwidth]{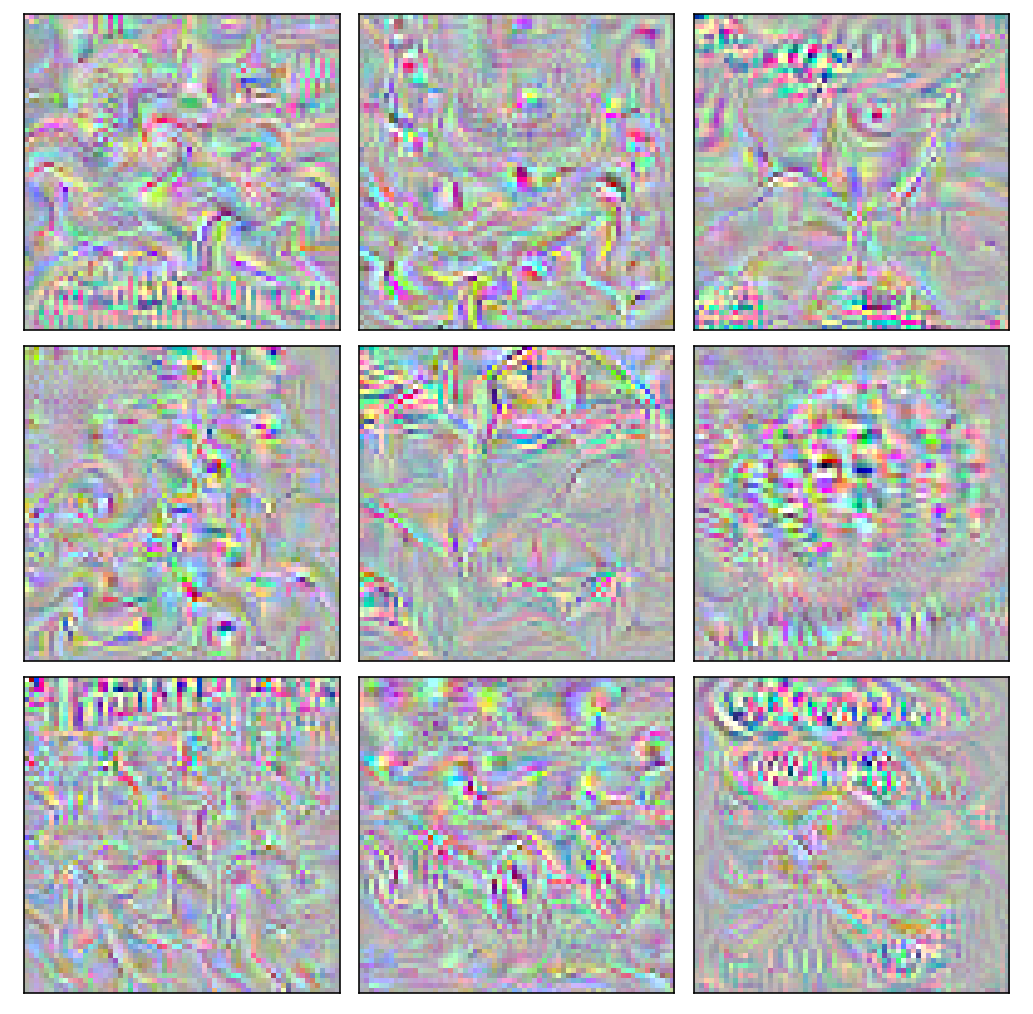}
        \caption{ResNet+edge, Tiny ImageNet}
        \label{fig:max-activation-edge-imagenet}
    \end{subfigure}    
    \caption{Activation maximization visualization of randomly selected units in the last convolutional layer.}
    \label{fig:max-activation}
\end{figure*}


\begin{figure*}[!bthp]
    \centering
    \includegraphics[width=0.5\textwidth]{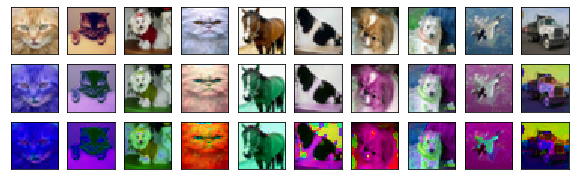}
    \caption{Some examples of color-shifted CIFAR-10 images with and without saturation regulation. First row: original images; second row: transformed images with the change saturation channel limited to avoid clipping; third row: transformed images with saturation delta sampled from [-1,1].}
    \label{fig:saturation-clipping}
\end{figure*}{}

\subsection{Activation Maximization Visualization}

To continue exploring the difference between models that differ in the ability to capture edges, we plot the max-activation visualization of two VGG models with and without edge-detection neurons with test accuracy 78.4\% and 83.2\%, respectively. For each neuron in the last convolutional layer of each model, we modify an $64 \times 64$ input images, starting from random noise, to maximize the activation of the neuron. Visualization is done using a modified version of Flashtorch \cite{Ogura.javin2020}.

Figure~\ref{fig:max-activation} shows the results of the above procedure for a random selection of neurons. Both CIFAR-10 models show circle-like and cone-like structures and similar alternation of high- and low-frequency regions. However, the visualization of the regular model (Figure~\ref{fig:max-activation-regular}) is more vibrant and contains bigger regions of solid colors. The model with edge-detection neurons (Figure~\ref{fig:max-activation-edge}) does display colors but with lower saturation and often in thin lines.
Similarly, the regular model trained on Tiny ImageNet uses a diverse set of colors and regions with solid or gradient filling (Figure~\ref{fig:max-activation-regular-imagenet}). On the other hand, the edge-detection model mostly generates lines and high-frequency color alternation.

Visualizations are a way to interpret models but without a theory about what a visualization should look like, it is hard to know what to look for and what predictions to make.
Based on the above observations, we hypothesize that neural networks that generate colorful visualizations are also more sensitive to color noise compared to those whose visualizations are less colorful.
Pointing in the same direction, it has been noticed that Inception \cite{szegedy2015}, a model that generates vibrant visualizations \cite{olah2020zoom}, also uses color as a discriminating feature~\cite{Hohman2020}. 
Similarly, \cite{Mu.andreas2020} reports that it is possible to flip the prediction of AlexNet, ResNet, and DenseNet by changing the color of a body of water from green to blue.

\subsection{Discussions}

\rev{The results in this section show that neural networks equipped with edge detection neurons are more robust to color distortion compared to conventional CNNs, 
but are also less accurate in natural settings.
The proposed models reach between 84\% and 99\% of the performance of conventional ones.
On the one hand, this shows the usefulness of the extracted color-independent features. On the other hand, it raises the question of the applicability of the new models to which we will give a two-pronged answer.
First, the i.i.d.\ datasets used in our experiments might contain spurious color cues that are accessible to regular CNNs only. It has been shown repeatedly that deep neural networks find shortcuts to solve a dataset without acquiring the intended capability \cite{Geirhos2020}. A regular CNN might acquire color detectors (green, blue, red), as demonstrated in  Figure~\ref{fig:weight-regular}, that are predictive of certain classes (frog/deer, bird/airplane/ship, car). Models with edge detection neurons are blocked from such features but their superior performance on transformed images suggest that the true capability is higher than what conventional benchmarks expose. Further testing on natural images in ``uncommon'' settings (i.e. ImageNet-A \cite{Hendrycks2019a}) will shed light on the true capability of the models.
Second, it is possible that edge detection neurons require a different architecture in higher layers to reach the high levels of performance reported in the literature.
CNN architectures with a standard first layer have been tuned on CIFAR-10 and ImageNet for many years; a new line of models will need further experimenting to reveal their best.
}

\rev{It has been suggested that achieving robustness in artificial intelligence amounts to acquiring so-called \textit{deep understanding} \cite{Marcus2020}. We argue that, in image recognition, this includes abstracting away from colors.
The concept of a car is independent of color but rather dependent on its shape, and humans can recognize a car with the same effortlessness whether it is of a rare color or a popular one.}
This is by no means an isolated case. Plants and animals, which constitute a large part of ImageNet, can be classified based on various morphological features but are rarely different by color only.
People with achromatopsia (total color blindness) can recognize everyday objects and function normally in society.
These observations suggest that shapes, which are composed of edges, are more fundamental properties than colors and disregarding colors brings us closer to a true understanding of concepts \rev{despite temporary setbacks in performance}.

Other research has pointed out that neural networks also demonstrate an over-reliance on superficial regularities \cite{Jo.bengio2017} and textual cues \cite{Geirhos2019}. 
We observe that the color-shifting procedure proposed in \cite{Hosseini.poovendran18} also causes structural distortion when the saturation channel is clipped (Figure~\ref{fig:saturation-clipping}).
This is because when saturation is changed unevenly, minor features might gain salience relative to others while salient features might become unrecognizable. Therefore, the drop of performance observed in \cite{Hosseini.poovendran18} should be interpreted as the reaction of neural networks to color \textit{and} structural noise.
In this paper, we have eliminated such structural distortion to focus on color.
However, it is possible that robust edge detection also plays a role in increasing structural robustness and shape bias in neural networks.

\section{Conclusions}
\label{sec:conclusions}

Our investigation into edge representations in neural networks has led to several insights. First, capturing edges robustly is harder than previously assumed: at least two layers are needed to detect edges independent of colors \rev{and they might not spontaneously acquire this capability from training on natural images}. 

Second, improving the robustness of edge detection in the early layers (i.e. the stability of their activation on color distortion) leads to higher robustness of the network against color-shifting and especially negative images.
The proposed edge detection units generate markedly different representations from regular neurons. 
We show that between 84\% and 99\% of performance on natural images can be achieved without accessing absolute color information.

\rev{Third, our analysis sheds light on how training on color-shifted images affects the representations of neural networks: it improves the accuracy of edge detection in lower layers but does not significantly improves their stability.}
Expanding on \cite{Hosseini.poovendran18}, we test on another architecture and dataset and, by means of visualization, point to the possible prevalence of the phenomena.

The current paper is only the first attempt at implementing edge detection neurons. We foresee that the research can be furthered along multiple directions. \rev{Different color spaces and normalization techniques can be applied to improve robustness.} Investigating how shapes can be captured in higher layers by combining edges is important to understand the internal representations of neural networks and to improve performance. \rev{Last but not least, experiments on ``uncommon'' or out of distribution images will bring valuable insights.}

{\small
\bibliographystyle{ieee_fullname}
\bibliography{Mendeley-Minh}
}

\end{document}